\documentclass[table]{acmart}
\AtBeginDocument{%
  \providecommand\BibTeX{{%
    \normalfont B\kern-0.5em{\scshape i\kern-0.25em b}\kern-0.8em\TeX}}}

\setcopyright{acmcopyright}
\copyrightyear{2018}
\acmYear{2018}
\acmDOI{XXXXXXX.XXXXXXX}

\acmConference[Conference acronym 'XX]{Make sure to enter the correct
  conference title from your rights confirmation emai}{June 03--05,
  2018}{Woodstock, NY}
\acmBooktitle{Woodstock '18: ACM Symposium on Neural Gaze Detection,
 June 03--05, 2018, Woodstock, NY} 
\acmPrice{15.00}
\acmISBN{978-1-4503-XXXX-X/18/06}

\usepackage{xcolor}
\usepackage{lipsum}
\usepackage{wasysym}
\usepackage{cleveref}
\usepackage{color,soul}
\usepackage{booktabs}
\usepackage{multirow}
\usepackage{makecell}
\usepackage{todonotes}
\usepackage[inline,shortlabels]{enumitem}

\usepackage{array}
\usepackage{longtable}

\makeatletter
\renewcommand{\noindentparagraph}{%
  \@startsection{paragraph}{4}%
  {\z@}{1.0ex \@plus 1ex \@minus .2ex}{-1em}%
  {\normalfont\normalsize\bfseries}%
}
\makeatother

\makeatletter
\newcommand{\crefnames}[3]{%
  \@for\next:=#1\do{%
    \expandafter\crefname\expandafter{\next}{#2}{#3}%
  }%
}
\makeatother

\crefnames{part,chapter,section}{\S}{\S\S}

\definecolor{lightgray}{gray}{0.95}
\let\oldtabular\tabular
\let\endoldtabular\endtabular
\renewenvironment{tabular}{\rowcolors{2}{white}{lightgray}\oldtabular}{\endoldtabular}
\newcommand{\ra}[1]{\renewcommand{\arraystretch}{#1}}

\begin{document}

\title{Trends in Explainable AI (XAI) Literature}


\author{Alon Jacovi}
\affiliation{%
  \institution{Bar Ilan University}
  \country{Israel}}
\email{alonjacovi@gmail}

% \renewcommand{\shortauthors}{}

\settopmatter{printacmref=false} % Removes citation information below abstract
\renewcommand\footnotetextcopyrightpermission[1]{} % removes footnote with conference information in first column
\pagestyle{plain} % removes running headers

\begin{abstract}

The XAI literature is decentralized, both in terminology and in publication venues, but recent years saw the community converge around keywords that make it possible to more reliably discover papers automatically. We use keyword search using the SemanticScholar API and manual curation to collect a well-formatted and reasonably comprehensive set of 5199 XAI papers, available at \url{https://github.com/alonjacovi/XAI-Scholar}. We use this collection to clarify and visualize trends about the size and scope of the literature, citation trends, cross-field trends, and collaboration trends. Overall, XAI is becoming increasingly multidisciplinary, with relative growth in papers belonging to increasingly diverse (non-CS) scientific fields, increasing cross-field collaborative authorship,  increasing cross-field citation activity. The collection can additionally be used as a paper discovery engine, by retrieving XAI literature which is cited according to specific constraints (for example, papers that are influential outside of their field, or influential to non-XAI research).

\end{abstract}

% \centering
%   \includegraphics[width=0.95\textwidth]{figures/teaser4.pdf}
%   \caption{Schematic of an explanatory narrative, as explored in this work. The narrative communicates a causal chain composed of two categories of causes: The objective causes in the context, and the actors' subjective interpretation of those causes. The causes' role is communicated against alternative contexts (contrast cases) that intervene on them. 
%   This paper develops the narrative structure, justifies it with precedence, and applies it to modern XAI methodologies to derive useful insights about the path towards successfully explaining complex AI.

\maketitle

\section{Introduction}

This is a report on the collection methodology and analysis of XAI-Scholar, a set of XAI\footnote{By XAI we refer to a relatively inclusive definition for research articles that discuss the development, implementation, or practice, of explanations/interpretations in ``AI'' systems (even if those articles don't refer to their work as explanations, or to their systems as AI systems, as long as they are treated as such by others). This definition is aligned with what we've observed in various curated lists, workshops and journal issues of XAI literature.} papers collected as of December 31st 2022. All data and reproduction code are available at \url{https://github.com/alonjacovi/XAI-Scholar}.

In recent years the “explainable AI” body of research has started to reach a size that makes it (1) difficult to grasp with manual surveying; (2) large enough that it’s possible to see overall empirical and statistical trends. The goal of this report is to collect a large and \textit{well-formatted} set of XAI papers to make this empirical analysis possible. The report mostly observes cross-field and multi-disciplinary trends in XAI. We invite others to use the collection for other purposes as well.   

\noindentparagraph{Challenges.}
XAI research has several properties that make it difficult to observe in its entirety, compared to many other adjacent fields in Computer Science:
\begin{enumerate}
    \item It is multidisciplinary, with non-negligible communities in many different fields that do not often interact or share venues.
    \item The terminology that papers use to self-identify as XAI research is not unique to XAI (E.g., “xai” and “Xai Xai” are names with multiple senses which appear in research), and this terminology is much more recent than the actual history of XAI.
    \item The most prevalent definitions of “XAI research” papers often include papers that don’t self-identify as XAI, as long as they research how to explain AI technology.
\end{enumerate}

\noindentparagraph{Findings.} Below is a brief partial summary of findings from the analysis. Most trends are situated around Computer Science (CS), being the primary field of study for XAI research. 
\begin{enumerate}
    \item XAI research has had its biggest ``expansion'' growth spikes outside of Computer Science in 2016, 2018 and 2021. 
    \item There is clear growth over time in the relative proportion of papers by authors that traditionally publish in two or more \textit{distinct} fields of study.
    \item CS has different citing relationships with different XAI fields. For example, XAI-CS cites XAI-Psychology more often than the vice versa, but the relationship flips for XAI-CS and XAI-Medicine. This ``direction'' of influence shows which fields often inform which fields in the current literature.
    \item There is a difference across XAI fields by how often they inform non-XAI research, the highest proportion being in XAI-Biology, XAI-Engineering and XAI-Law---while the lowest proportion being in XAI-Psychology, XAI-Business and XAI-Philosophy, whose influence more often carries to other XAI literature.
    \item Citation behavior across papers is significantly different between fields. For example, the top-cited Philosophy papers cited by XAI-Philosophy are significantly different from those cited by XAI-CS, and so on. Unsurprisingly, papers outside of a field tend to focus on a smaller variety of papers in that field, but the papers that ``break out of'' the traditional boundaries of their field are not always the most cited papers in that field.
    \item The collection can serve as a paper discovery engine by observing which XAI papers, for example, are the most influential to papers outside of their field, or outside of XAI; or which non-XAI papers of a particular field are the most informative to another field. Tables with examples of these selections are shown at the end of this report, and more are available in the accompanying github repository.
\end{enumerate}

\section{Data Collection}

\subsection{Methodology}
The collection phase involved five steps.

\noindentparagraph{Step 1: Keyword-based search (3101 papers total).}
We derive a set of keywords in a (manual) iterative process to maximize recall while not compromising near-perfect precision. The keywords are matched against both the title and abstract together. Due to problems with filtering papers with 1 keyword match, we only collect papers that match 2 keywords or more. A random sample of 100 papers yielded 99\% precision. 

The keywords are: \textit{xai, (xai), hcxai, explainability, interpretability,  explainable ai, explainable artificial intelligence, interpretable ml, interpretable machine learning, interpretable model, feature attribution, feature importance, global explanation, local explanation, local interpretation, global interpretation, model explanation, model interpretation, saliency, counterfactual explanation.}

\noindentparagraph{Step 2: Manually curated collections (+ 766; 3867 papers total).}
We collected the titles of papers from various curated XAI collections \cite{danilevsky2020survey,mohseni2018multidisciplinary,s1,s2,s3,s4,s5,s6,s7,s8,s9,s10,s11} and matched them using the SemanticScholar API with fuzzy matching.

\noindentparagraph{Step 3: Citation tree expansion with manual filtering (+ 648; 4515 papers total).} We took the 2000 most cited papers by the set of papers collected in the previous steps, and manually selected XAI papers from them.

\noindentparagraph{Step 4: Citation tree expansion with automatic filtering (+ 709; 5224 papers total).} We used the citations and references of all collected papers and filtered them via the 2-keyword-match method from step 1. We repeated this until no new papers were added. 

\noindentparagraph{Step 5: Manual quality check (- 25; 5119 papers total).} Finally, we heuristically found 25 incorrectly-attributed papers in the set, which we removed from the collection.

\newpage

\subsection{Collection Details}

\begin{figure*}[ht]
  \includegraphics[width=0.99\textwidth]{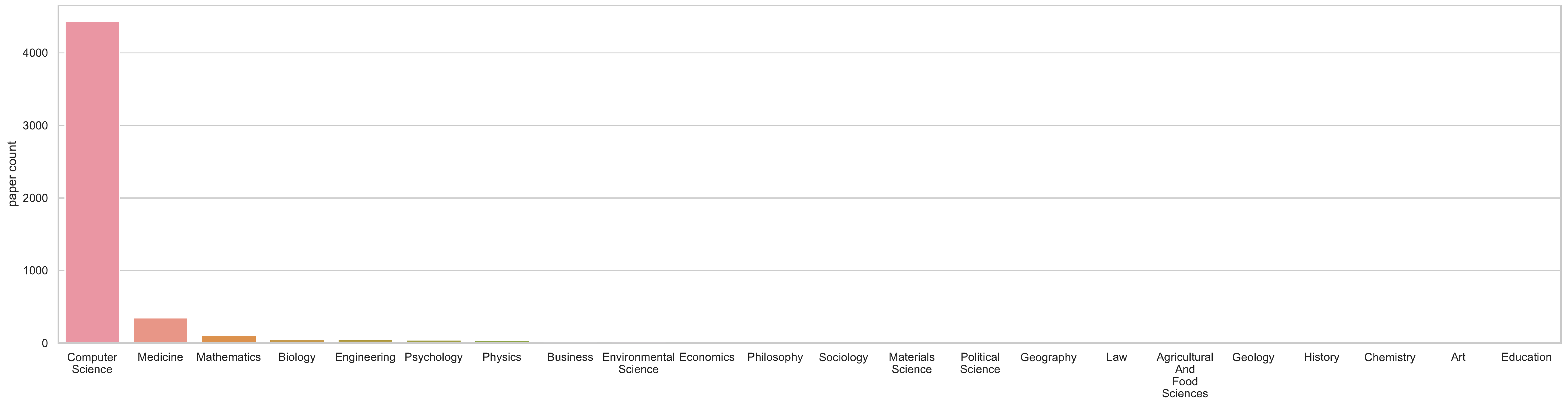}
  \includegraphics[width=0.99\textwidth]{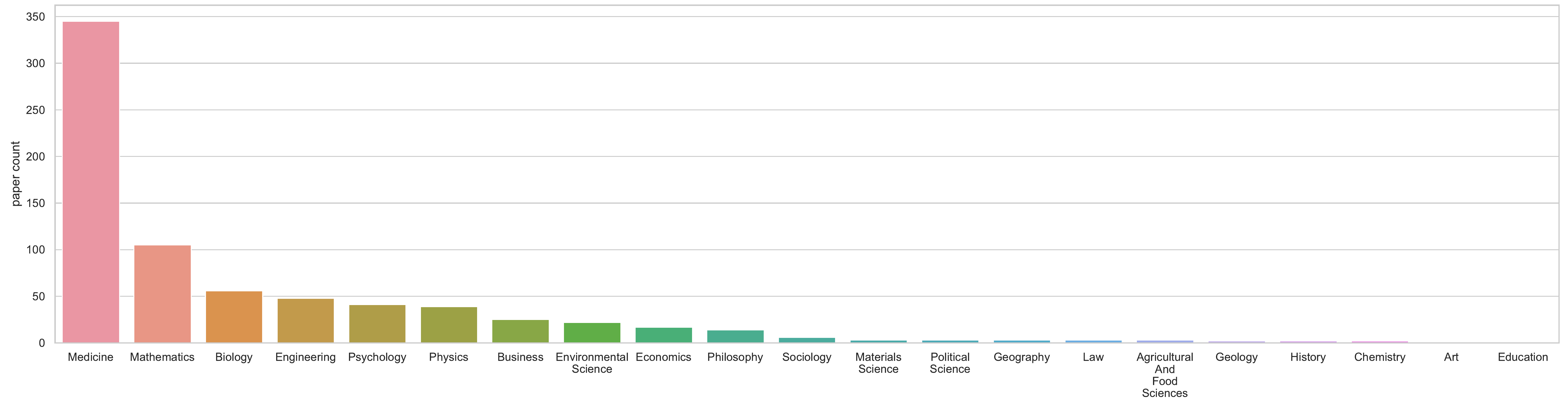}
  \caption{Paper counts by field of study. The bottom figure shows distribution across fields, sans Computer Science, for readability.}
  \label{fig:field-counts}
\end{figure*}

The collection has 5199 papers. Each paper in the collection contains, per the SemanticScholar API:
\begin{enumerate}
    \item The SemanticScholar ID and URL
    \item Title
    \item Abstract
    \item Authors
    \item Number of citations 
    \item Number of references
    \item Year
    \item Venue
    \item Field of study
    \item SemanticScholar’s “tldr” summary
        \item References
    \item Citations
    \item SemanticScholar's embedding vector
\end{enumerate}

See \Cref{fig:field-counts} for field of study distributions.

\newpage

\noindentparagraph{Limitations.} Any insight derived from the collection should account for margin of error based on these limitations:
\begin{enumerate}
    \item The data in the collection is noisy as can be expected from a large-scale research database. This includes missing fields, inconsistent venue names, some incorrect details, and so on. While this is the minority, it is not negligible.
    \item It's likely that the collection still contains a very small amount of non-XAI papers, despite our efforts.
    \item The collection methodology is biased towards CS, influential papers and papers that self-identify as XAI with the keywords that we used. Of course, it's likely that many XAI papers were missed, in particular less-cited papers and papers which use different terminology.
    \item The collection was retrieved over a period of time in December 2022. Due to delay in proceedings release for some venues, it may be necessary to consider 2022 a partial year. 
\end{enumerate}

\section{Growth Trends}

\begin{figure*}[ht]
  \includegraphics[width=0.82\textwidth]{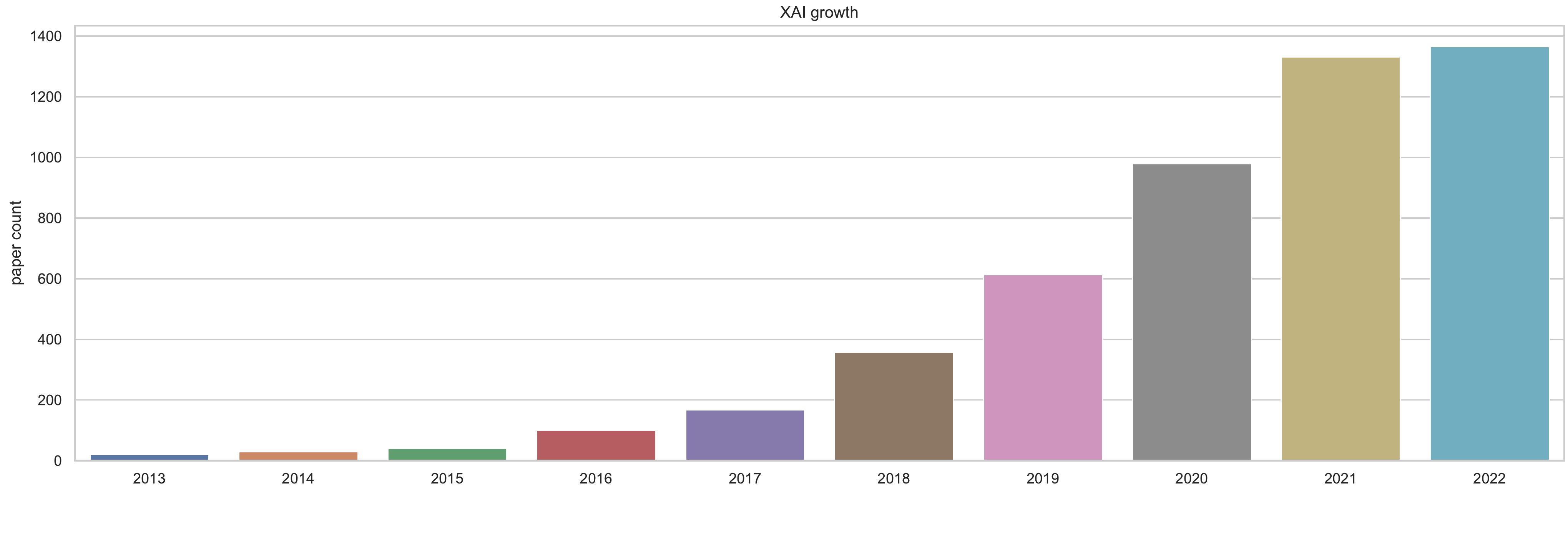}
  \includegraphics[width=0.82\textwidth]{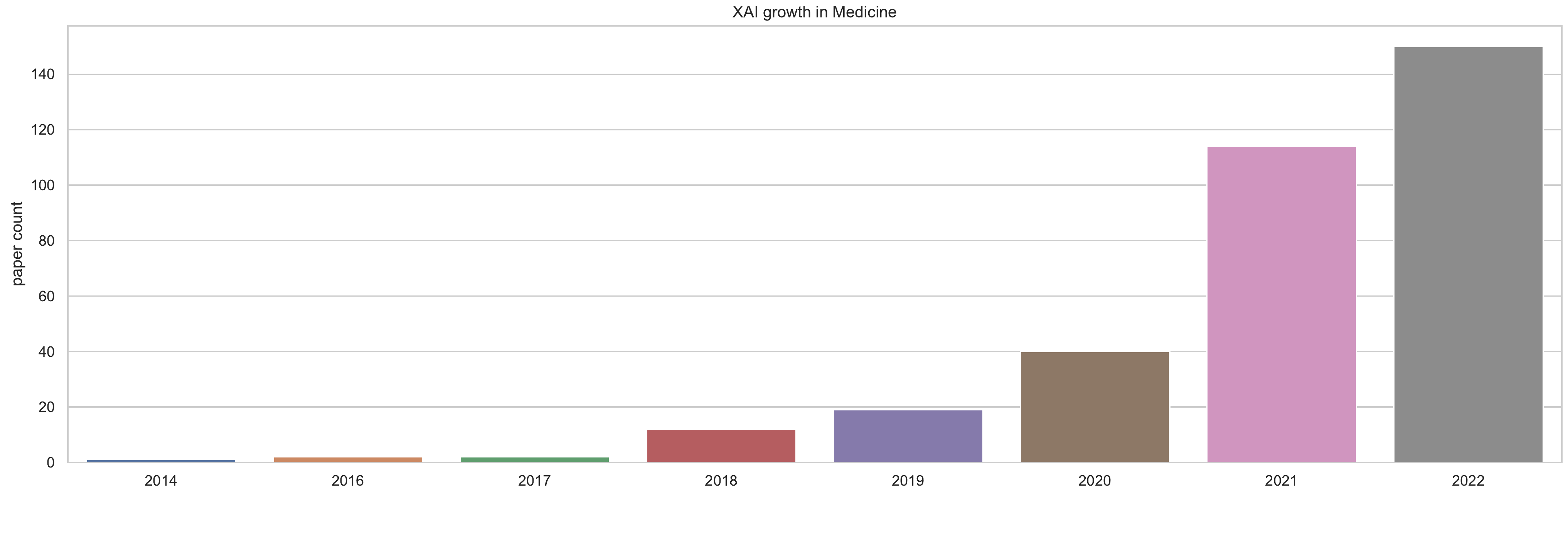}
  \includegraphics[width=0.82\textwidth]{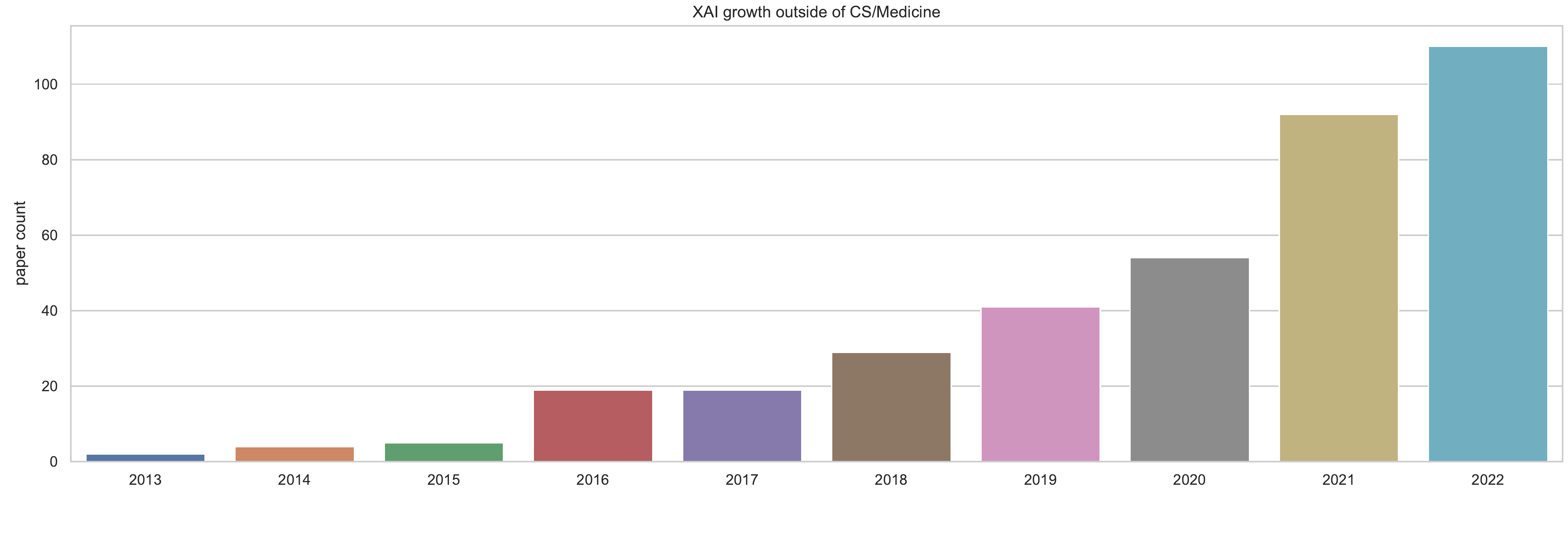}
  \caption{Yearly growth trends. Papers before 2013 were omitted for readability.}
  \label{fig:yearly-growth}
\end{figure*}

\Cref{fig:yearly-growth} shows yearly growth in three settings. First, XAI generally shows relative yearly growth---but this growth is largely controlled by Computer Science (which shows the same growth trend). 

Controlling for non-CS papers reveals different growth trends. For example, XAI-Medicine shows exponential growth, in particular with large relative growth in 2016, 2018 and 2021. These trends also hold when controlling for non-Medicine and non-CS papers (bottom plot). Overall, it appears that XAI has had the biggest growth into \textit{non-central fields} in 2016, 2018 and 2021.

\section{Collaboration Trends}

We define the field of study for an author as the field of study for a majority of their papers (retrieved separately via the SemanticScholar API). Given this definition, we can define a ``collaboration'' as the existence of two authors for the same paper assigned two different fields of study.

\begin{figure*}[ht]
  \includegraphics[width=0.49\textwidth]{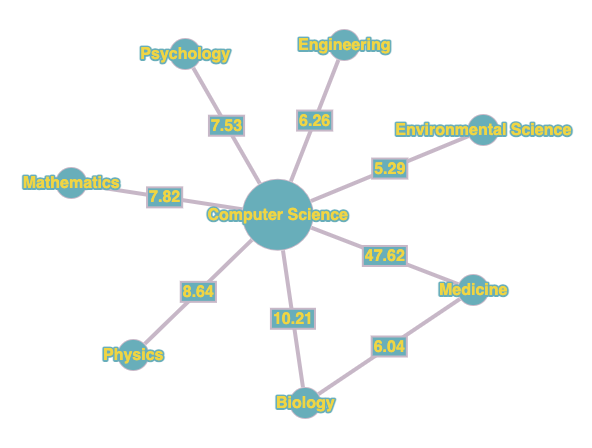}
  \includegraphics[width=0.49\textwidth]{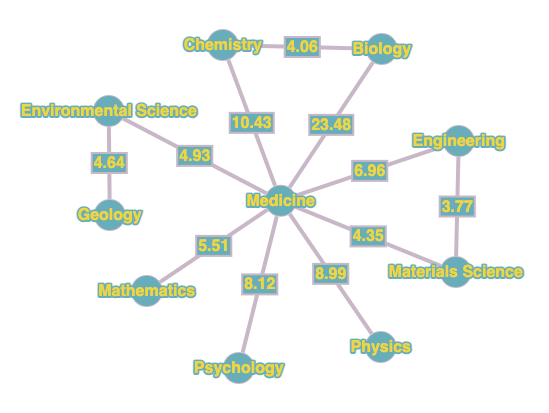}
  \caption{Cross-field collaboration as a weighted undirected graph for all field pairs (left) and excluding CS (right). Edge weights show the percentage of collaborations for a field pair out of all papers with any collaboration. Low-magnitude edges were omitted.}
  \label{fig:collab-graph}
\end{figure*}

\Cref{fig:collab-graph} shows the most common field pairings. As CS controls the scale, we show a graph that omits CS for a perspective on multi-field research between non-CS fields.  

\begin{figure*}[ht]
  \includegraphics[width=0.49\textwidth]{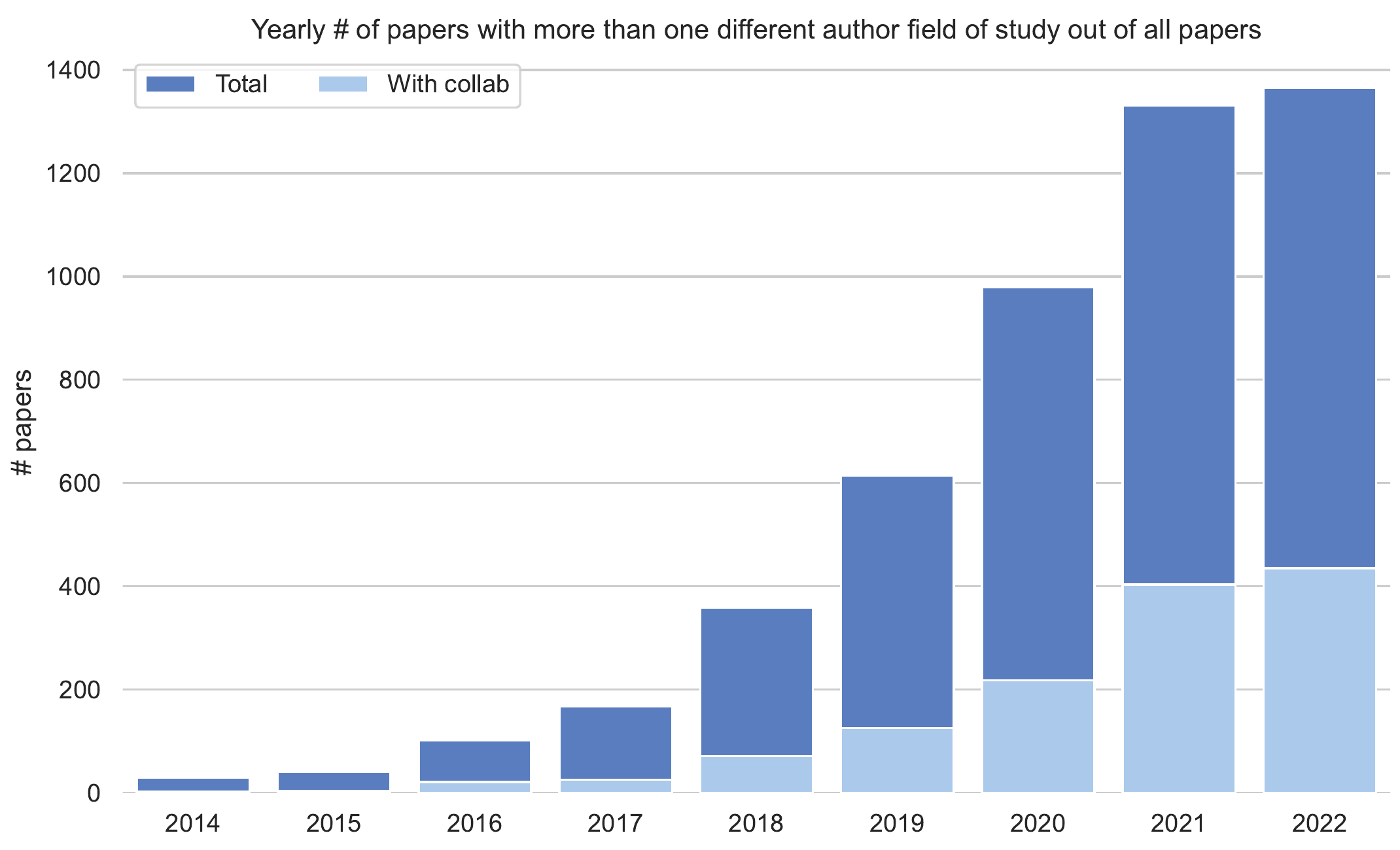}
  \includegraphics[width=0.49\textwidth]{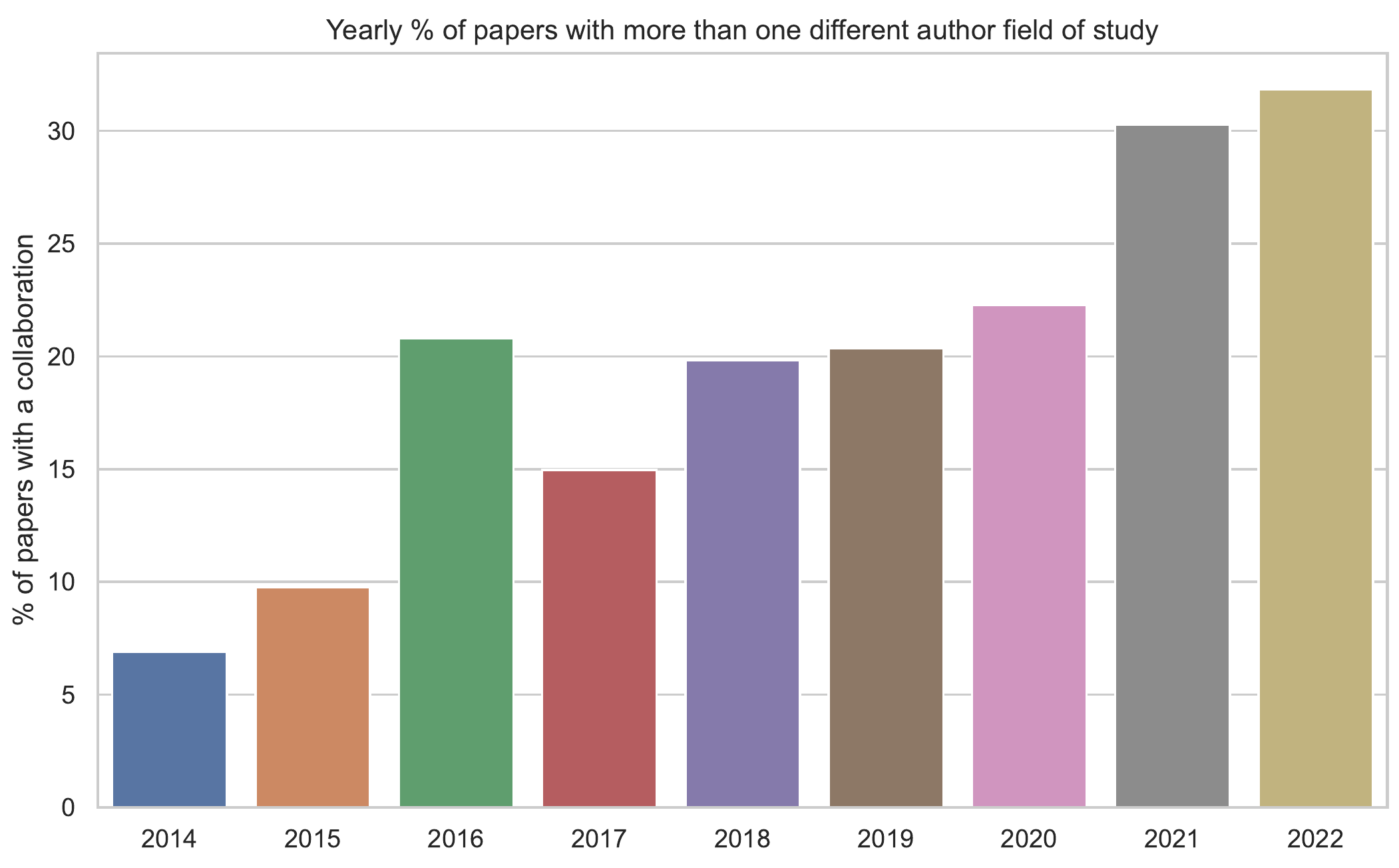}
  \caption{Absolute and relative plots of yearly papers with authors of at least two fields of study, out of all papers.}
  \label{fig:collab-plot}
\end{figure*}

\Cref{fig:collab-plot} shows yearly absolute and relative growth for papers with at at least two different fields, via their authors, out of all yearly papers. If we consider 2016 as an outlier, the trends show a clear relative growth in cross-field XAI research.

\section{Citation Trends}

This section investigates the interaction between the citation graph in XAI-Scholar and the fields of study variable.

\subsection{Computer Science}

\begin{figure}[ht]
  \includegraphics[width=0.6\textwidth]{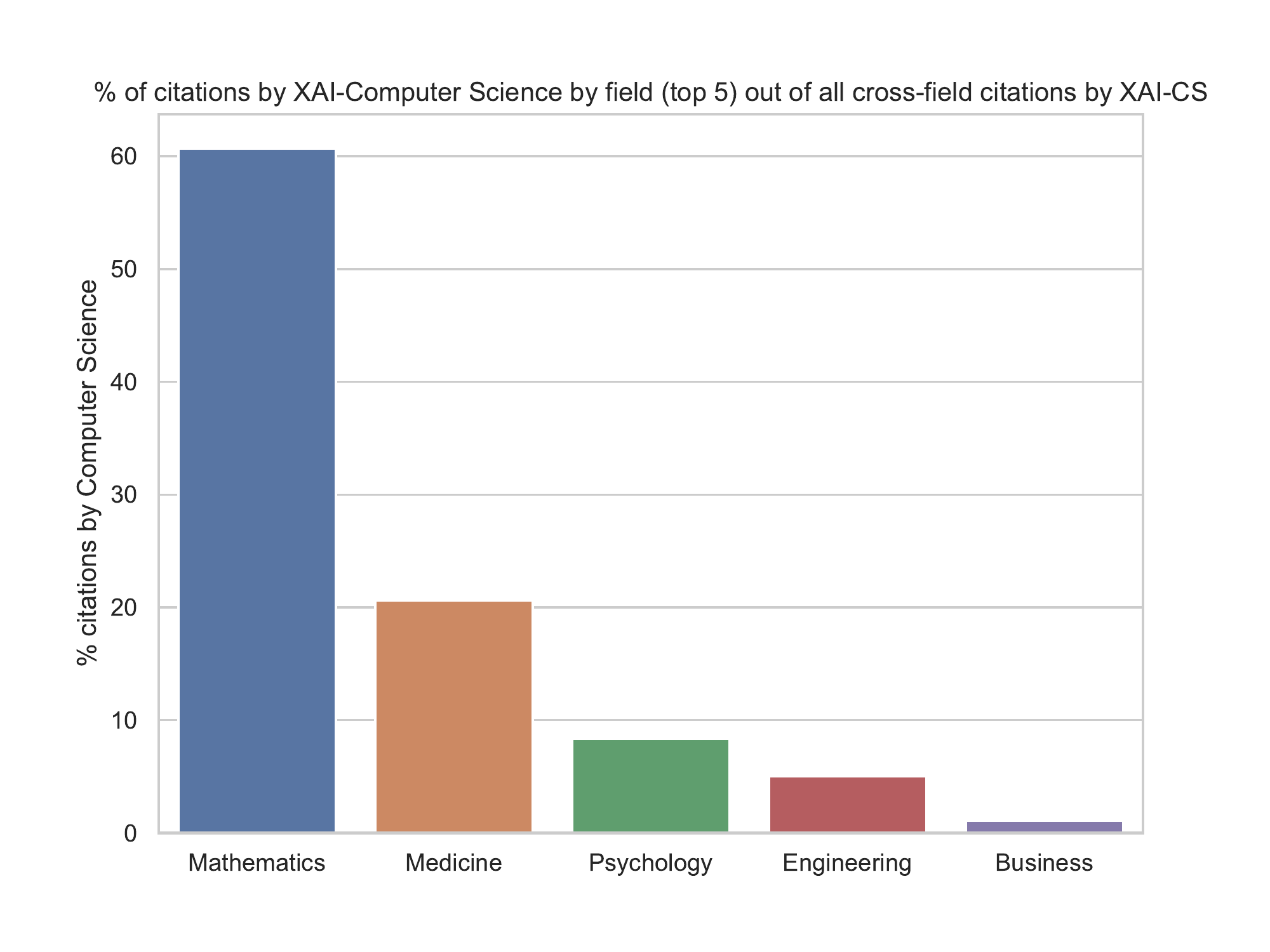}
  \caption{The top-5 XAI fields cited by XAI-CS.}
  \label{fig:cited-by-cs}
\end{figure}

\Cref{fig:cited-by-cs} shows the distribution of citations by field by XAI-CS literature. Unsurprisingly XAI-CS appears to be significantly informed by XAI-Mathematics, but XAI-Medicine, XAI-Psychology, XAI-Engineering and XAI-Business also have a non-negligible presence. 

\begin{figure*}[ht]
  \includegraphics[width=0.70\textwidth]{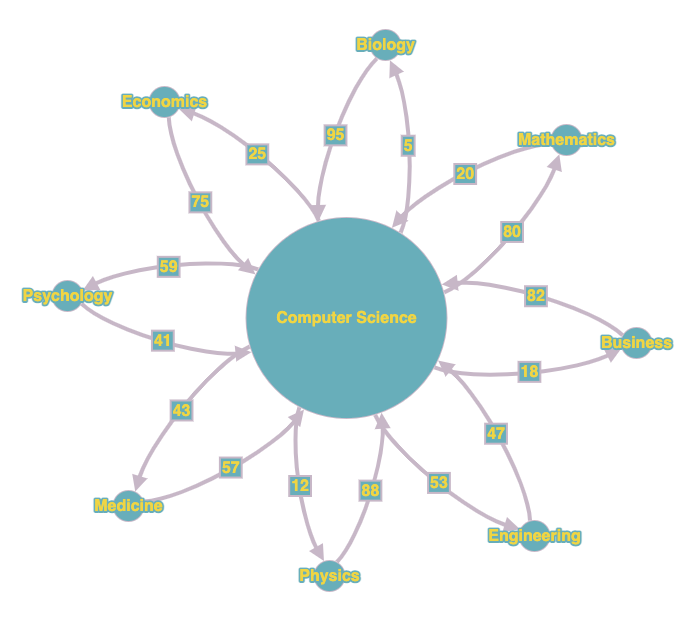}
  \caption{Citation relationship between XAI-CS and other XAI fields. Every pair of edges is normalized to 100\% to show which of the two fields is cited more by the other.}
  \label{fig:cs-relationships}
\end{figure*}

\Cref{fig:cs-relationships} shows the citation relationship between XAI-CS and other XAI fields as a directed weighted graph. Every pair of edges is normalized to 100\% to show which of the two XAI fields is informed more by the other. We can observe that XAI-CS more often cites (vs. is cited by) XAI-Psychology and XAI-Mathematics, while the same is not true for the other fields (XAI-Engineering and XAI-Engineering being roughly equal, and the rest overwhelmingly citing XAI-CS).

\subsection{Citation Trends Outside of Computer Science}

\begin{figure*}[ht]
  \includegraphics[width=0.70\textwidth]{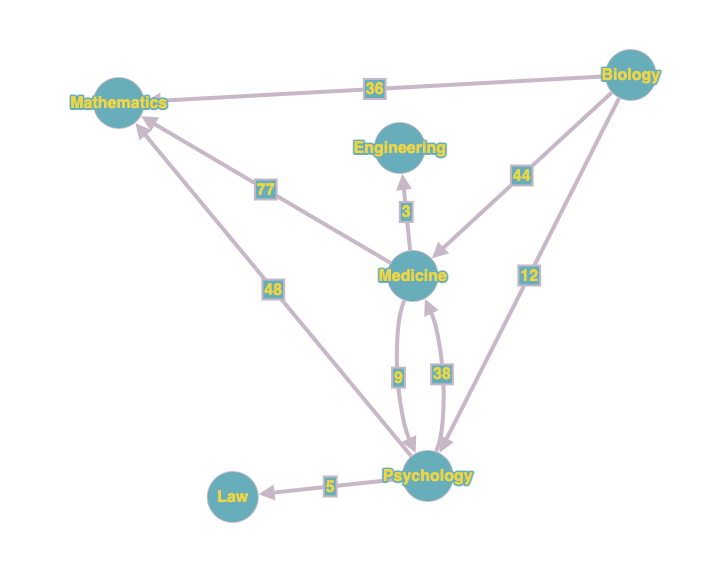}
  \caption{A graph of the top-3 cited XAI fields for every XAI field (omitting fields whose outgoing citations are less than 20). All outgoing edges from a node are normalized to sum to 100\%.}
  \label{fig:noncs-relationships}
\end{figure*}

As before we can omit XAI-CS in order to probe into relationships between other fields. \Cref{fig:noncs-relationships} shows a cross-field citation graph where all outgoing edges from a node are normalized to sum to 100\%, to show which XAI fields are most cited by a particular XAI field.

\subsection{Citation Trends Between XAI and non-XAI}

\begin{figure*}[ht]
  \includegraphics[width=0.85\textwidth]{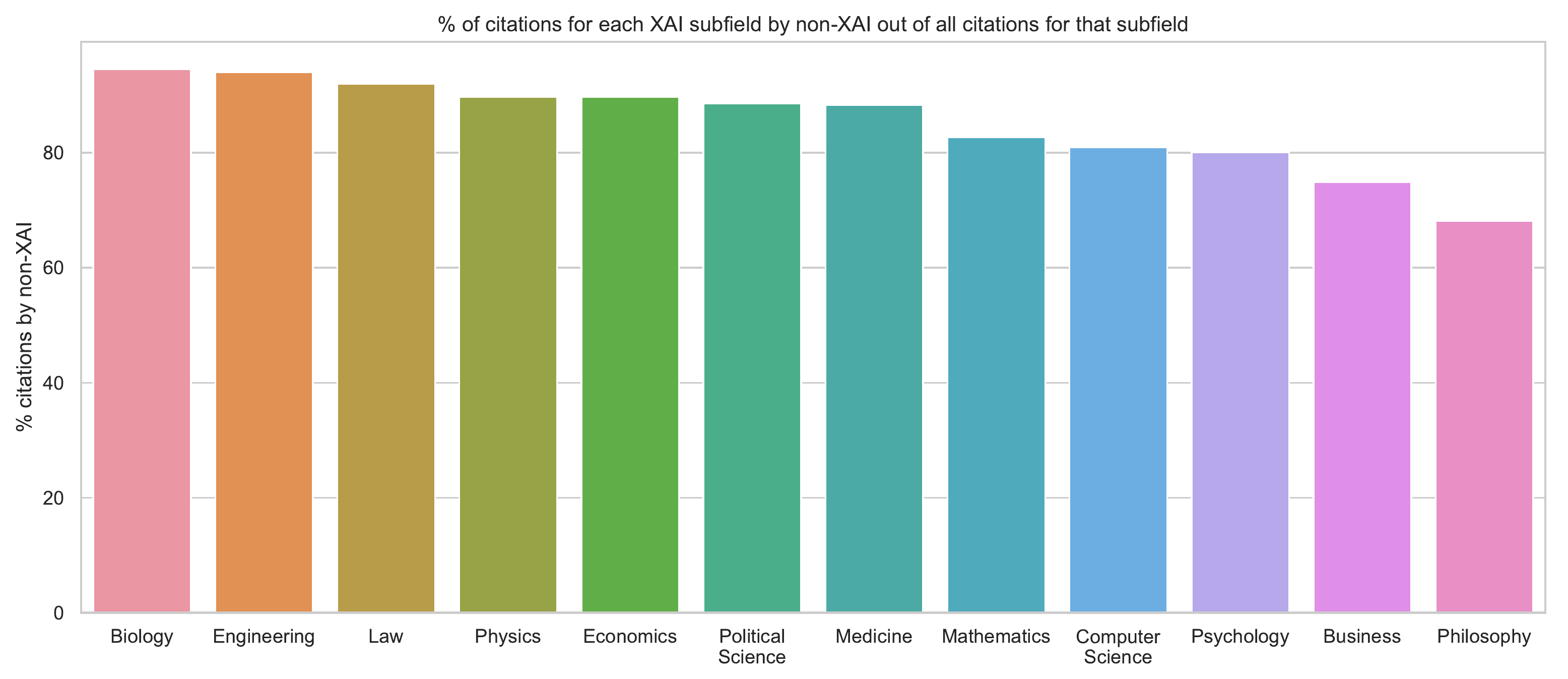}
  \caption{The percentage of non-XAI citations (out of all citations) per XAI field. Fields with less than 100 outgoing citations were omitted.}
  \label{fig:nonxai-citation}
\end{figure*}

\Cref{fig:nonxai-citation} show the ranking of XAI fields by the percentage of non-XAI citations out of all citations to that XAI field. This plot shows which fields most often inform non-XAI literature. For example, XAI-Biology is relatively often cited by non-XAI literature, and the opposite is true for XAI-Philosophy, which seems to be comparatively more often cited by XAI.

% \newpage

\section{Paper-level Citation Trends and Paper Discovery}

\begin{figure*}[ht]
  \includegraphics[width=0.99\textwidth]{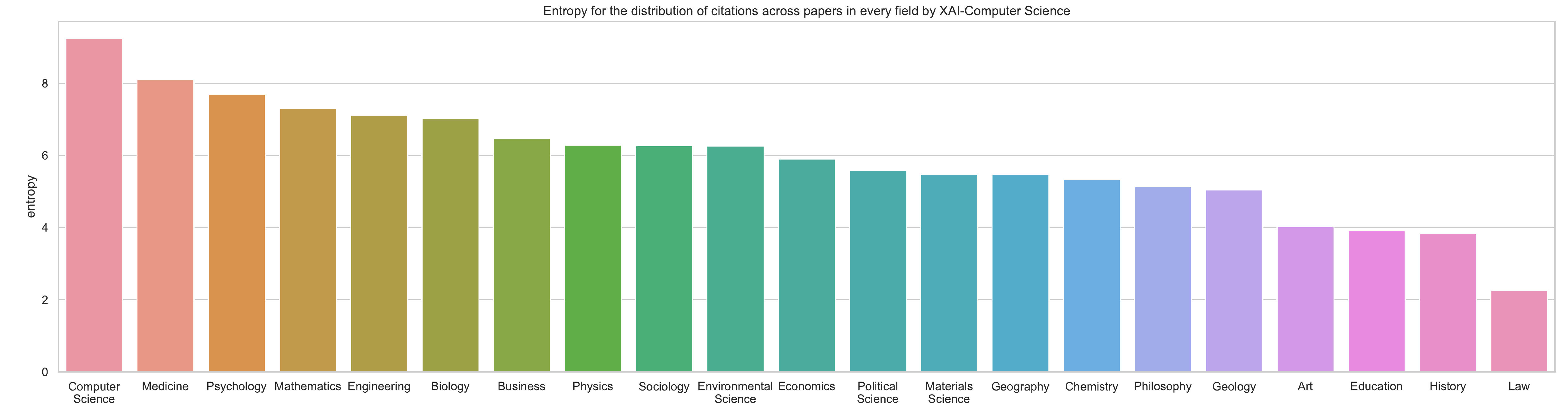}
  \caption{Plot of the entropies for every distribution of citations by XAI-CS across papers, by cited field. For example, XAI-CS citations of Law papers are concentrated in fewer papers controlling more of the citations, compared to citations of Biology papers.}
  \label{fig:entropy}
\end{figure*}

Finally we can look at citation behavior at the level of individual papers. For example, when XAI-CS cites Philosophy, are they citing a wide variety of papers, or a select minority of papers? \Cref{tab:phil2xai-cs} shows the top cited papers in Philosophy by XAI-CS, which control 32\% of the citations between these fields. More generally, \Cref{fig:entropy} shows the entropy across paper citations for every field as cited by XAI-CS. The field with the most extreme distribution is Law, which as shown in \Cref{tab:law2xai-cs}, is due to one paper---``Accountable Algorithms''---being overwhelmingly the most cited paper in Law by XAI-CS. \Cref{tab:psy2xai-cs} and \Cref{tab:cs2xai-cs} show additional examples for top papers in Psychology and CS as cited by XAI-CS.

Other discovery constraints include: 
\begin{enumerate}
    \item Which papers in a particular field are the most cited by XAI papers of that field? (e.g., \Cref{tab:math2xai-math,tab:psy2xai-psy})
    \item What are the XAI papers that are the most cited by papers \textit{outside} of their field? (e.g., \Cref{tab:xai-cs2non-cs,tab:xai-eco2non-eco,tab:xai-bio2non-bio})
    \item What XAI papers in a particular field are most cited in that field?
    \item And so on.
\end{enumerate}

\section{Conclusions}

XAI research is converging around specific terminology at scale that makes it possible to observe trends empirically. While the retrieval process has some limitations that result in a margin of error, it's possible to account for them on some level (for example, by acknowledging that the retrieval is biased towards Computer Science and seeing trends that overcome this bias in the opposite direction). The analysis in this work mostly focused on the field of study variable, though it is possible to look at many other trends given this collection, as has been explored for other fields \cite{citat}.

\begin{acks}
Thanks to Matan Eyal, Mor Geva, Avi Caciularu, Yoav Goldberg, Yonatan Bitton and Rotem Tsabary, for discussions and brainstorming.
\end{acks}

\begin{table}[ht]
\ra{1.3}
    \centering
\resizebox{0.7\linewidth}{!}{
    \begin{tabular}{c l}
    \toprule
        \% cited & Title \\
        \midrule
        6.4\%	&Contrastive Explanation \\
5.6\%	&Causality: Models, Reasoning and Inference \\
3.5\%	&The Book of Why: The New Science of Cause and Effect \\
3.5\%	&Studies in the Logic of Explanation \\
3.1\%	&Causality \\
3.1\%	&The philosophical basis of algorithmic recourse \\
2.0\%	&Scientific Explanation and the Causal Structure of the World \\
1.9\%	&Scientific Explanation \\ 
1.7\%	&Knowledge-Based Causal Attribution : The Abnormal Conditions Focus Model \\
1.2\%	&Combining explanation and argumentation in dialogue \\
\bottomrule
    \end{tabular}}
    \caption{Top cited papers in Philosophy by XAI-CS.}
    \label{tab:phil2xai-cs}
\end{table}

\begin{table}[ht]
\ra{1.3}
    \centering
\resizebox{0.6\linewidth}{!}{
    \begin{tabular}{c l}
    \toprule
        \% cited & Title \\
        \midrule
44.4\%&Accountable Algorithms\\
7.4\%&Stability\\
7.4\%&The Role of Explanation in Algorithmic Trust\\
3.7\%&HYPO’S legacy: introduction to the virtual special issue\\
3.7\%&THE UNIVERSITY OF CHICAGO LAW REVIEW\\
3.7\%&The Philadelphia predictive policing experiment\\
3.7\%&Race , Prediction , and Discretion\\
1.9\%&Antidiscriminatory Algorithms\\
1.9\%&Of, for, and by the people: the legal lacuna of synthetic persons\\
1.9\%&FORECASTING THE FUTURE OF PREDICTIVE CRIME MAPPING\\
\bottomrule
    \end{tabular}}
    \caption{Top cited papers in Law by XAI-CS.}
    \label{tab:law2xai-cs}
\end{table}

\begin{table}[ht]
\ra{1.3}
    \centering
    
\resizebox{0.8\linewidth}{!}{
\begin{tabular}{c p{12cm}}
    \toprule
        \% cited & Title \\
        \midrule
1.0\%&The structure and function of explanations\\
1.0\%&The Role of Explanations on Trust and Reliance in Clinical Decision Support Systems\\
0.8\%&Explanation and understanding.\\
0.7\%&False Positives, False Negatives, and False Analyses: A Rejoinder to "Machine Bias: There's Software Used across the Country to Predict Future Criminals. and It's Biased against Blacks"\\
0.6\%&A unified view of gradient-based attribution methods for Deep Neural Networks\\
0.6\%&Conversational Processes and Causal Explanation\\
0.6\%&When Explanations Lie: Why Many Modified BP Attributions Fail\\
0.5\%&Explanation and Abductive Inference\\
0.4\%&Causal Inference in Statistics: A Primer\\
0.4\%&Humans and Automation: Use, Misuse, Disuse, Abuse\\
\bottomrule
    \end{tabular}}
    \caption{Top cited papers in Psychology by XAI-CS.}
    \label{tab:psy2xai-cs}
\end{table}

% \begin{table}[ht]
% \ra{1.3}
%     \centering
% \resizebox{0.9\linewidth}{!}{
% \begin{tabular}{c p{10cm}}
%     \toprule
%         \% cited & Title \\
%         \midrule
% 3.8\%&The philosophy of artificial intelligence\\
% 3.8\%&Models and Modeling in the Sciences\\
% 3.8\%&The philosophical basis of algorithmic recourse\\
% 3.8\%&Idealizations and Understanding: Much Ado About Nothing?\\
% 3.8\%&What Is Understanding? An Overview of Recent Debates in Epistemology and Philosophy of Science\\
% 3.8\%&Explaining Understanding : New Perspectives from Epistemology and Philosophy of Science\\
% 3.8\%&How Idealizations Provide Understanding\\
% 3.8\%&The Ontic Account of Scientific Explanation\\
% 3.8\%&Idealizations and scientific understanding\\
% 3.8\%&Scientific Representation\\
% \bottomrule
%     \end{tabular}}
%     \caption{Top cited papers in Philosophy by XAI-Philosophy.}
%     \label{tab:psy2xai-cs}
% \end{table}

\begin{table}[ht]
\ra{1.3}
    \centering
\resizebox{0.9\linewidth}{!}{
    \begin{tabular}{c l}
    \toprule
        \% cited & Title \\
        \midrule
1.4\%&“Why Should I Trust You?”: Explaining the Predictions of Any Classifier\\
1.0\%&A Unified Approach to Interpreting Model Predictions\\
0.6\%&Deep Inside Convolutional Networks: Visualising Image Classification Models and Saliency Maps\\
0.6\%&Grad-CAM: Visual Explanations from Deep Networks via Gradient-Based Localization\\
0.6\%&Axiomatic Attribution for Deep Networks\\
0.5\%&Visualizing and Understanding Convolutional Networks\\
0.5\%&Towards A Rigorous Science of Interpretable Machine Learning\\
0.5\%&On Pixel-Wise Explanations for Non-Linear Classifier Decisions by Layer-Wise Relevance Propagation\\
0.4\%&Explanation in Artificial Intelligence: Insights from the Social Sciences\\
0.4\%&Learning Important Features Through Propagating Activation Differences\\
\bottomrule
    \end{tabular}}
    \caption{Top cited papers in CS by XAI-CS.}
    \label{tab:cs2xai-cs}
\end{table}

\begin{table}[ht]
\ra{1.3}
    \centering 
\resizebox{0.9\linewidth}{!}{
\begin{tabular}{c p{15cm}}
    \toprule
        \% cited & Title \\
        \midrule
1.4\%&Greedy function approximation: A gradient boosting machine.\\
1.1\%&Regression Shrinkage and Selection via the Lasso\\
1.0\%&European Union Regulations on Algorithmic Decision-Making and a "Right to Explanation"\\
1.0\%&Interpretability Beyond Feature Attribution: Quantitative Testing with Concept Activation Vectors (TCAV)\\
0.8\%&Peeking Inside the Black Box: Visualizing Statistical Learning With Plots of Individual Conditional Expectation\\
0.8\%&An Efficient Explanation of Individual Classifications using Game Theory\\
0.7\%&Understanding Black-box Predictions via Influence Functions\\
0.7\%&PREDICTIVE LEARNING VIA RULE ENSEMBLES\\
0.6\%&Equality of Opportunity in Supervised Learning\\
0.6\%&Generalized Functional ANOVA Diagnostics for High-Dimensional Functions of Dependent Variables\\
\bottomrule
    \end{tabular}}
    \caption{Top cited papers in Mathematics by XAI-Mathematics.}
    \label{tab:math2xai-math}
\end{table}

\begin{table}[ht]
\ra{1.3}
    \centering 
\resizebox{0.9\linewidth}{!}{
\begin{tabular}{c p{15cm}}
    \toprule
        \% cited & Title \\
        \midrule
1.7\%&Explanation and understanding.\\
1.0\%&How the Mind Explains Behavior: Folk Explanations, Meaning, and Social Interaction\\
1.0\%&Mental Models and Causal Explanation: Judgements of Probable Cause and Explanatory Relevance\\
1.0\%&Functional explanation and the function of explanation\\
1.0\%&The structure and function of explanations\\
1.0\%&The misunderstood limits of folk science: an illusion of explanatory depth\\
1.0\%&Explanatory coherence in social explanations : a parallel distributed processing account\\
1.0\%&Explanatory coherence\\
0.7\%&Attribution theory in social psychology\\
0.7\%&A unified view of gradient-based attribution methods for Deep Neural Networks\\
\bottomrule
    \end{tabular}}
    \caption{Top cited papers in Psychology by XAI-Psychology.}
    \label{tab:psy2xai-psy}
\end{table}

\begin{table}[ht]
\ra{1.3}
    \centering 
\resizebox{0.9\linewidth}{!}{
\begin{tabular}{c p{15cm}}
    \toprule
        \% cited & Title \\
        \midrule
1.9\%&Detection of Influential Observation in Linear Regression\\
1.8\%&Conditional variable importance for random forests\\
1.7\%&To Explain or to Predict\\
1.7\%&Bias in random forest variable importance measures: Illustrations, sources and a solution\\
1.5\%&Illuminating the “black box”: a randomization approach for understanding variable contributions in artificial neural networks\\
1.4\%&From local explanations to global understanding with explainable AI for trees\\
1.4\%&On the interpretation of weight vectors of linear models in multivariate neuroimaging\\
1.4\%&An accurate comparison of methods for quantifying variable importance in artificial neural networks using simulated data\\
1.3\%&Permutation importance: a corrected feature importance measure\\
1.2\%&How the machine ‘thinks’: Understanding opacity in machine learning algorithms\\
\bottomrule
    \end{tabular}}
    \caption{Top cited papers in XAI-CS by papers outside of CS.}
    \label{tab:xai-cs2non-cs}
\end{table}

\begin{table}[ht]
\ra{1.3}
    \centering 
\resizebox{0.9\linewidth}{!}{
\begin{tabular}{c p{15cm}}
    \toprule
        \% cited & Title \\
        \midrule
51.5\%&The Right to Explanation, Explained\\
37.2\%&The Shapley value\\
4.6\%&Economic complexity unfolded: Interpretable model for the productive structure of economies\\
2.5\%&Predicting , explaining , and understanding risk of long-term unemployment\\
1.3\%&Explainable AI Models of Stock Crashes: A Machine-Learning Explanation of the Covid March 2020 Equity Meltdown\\
1.3\%&Paving the way towards counterfactual generation in argumentative conversational agents\\
0.4\%&Explainable AI (XAI) Models Applied to Planning in Financial Markets\\
0.4\%&Sell Me the Blackbox! Why eXplainable Artificial Intelligence (XAI) May Hurt Customers\\
0.4\%&An Investigation of the Impact of COVID-19 Non-Pharmaceutical Interventions and Economic Support Policies on Foreign Exchange Markets with Explainable AI Techniques\\
0.4\%&Frontiers in environmental science a study on China coal price forecasting based on CEEMDAN-GWO-CatBoost hybrid forecasting model under carbon neutral target\\
\bottomrule
    \end{tabular}}
    \caption{Top cited papers in XAI-Economics by papers outside of Economics.}
    \label{tab:xai-eco2non-eco}
\end{table}

\begin{table}[ht]
\ra{1.3}
    \centering 
\resizebox{0.95\linewidth}{!}{
\begin{tabular}{c p{16cm}}
    \toprule
        \% cited & Title \\
        \midrule
28.9\%&Development and interpretation of a pathomics-based model for the prediction of microsatellite instability in Colorectal Cancer\\
20.5\%&Discovering epistatic feature interactions from neural network models of regulatory DNA sequences\\
8.4\%&Amino Acid k-mer Feature Extraction for Quantitative Antimicrobial Resistance (AMR) Prediction by Machine Learning and Model Interpretation for Biological Insights\\
6.3\%&Brain age prediction of healthy subjects on anatomic MRI with deep learning : going beyond with an “explainable AI” mindset\\
5.8\%&Reverse-engineering Recurrent Neural Network solutions to a hierarchical inference task for mice\\
5.8\%&Inferring Sequence-Structure Preferences of RNA-Binding Proteins with Convolutional Residual Networks\\
4.2\%&Automated detection of glaucoma with interpretable machine learning using clinical data and multi-modal retinal images\\
3.2\%&BayeSuites: An open web framework for massive Bayesian networks focused on neuroscience\\
3.2\%&Analysis of SARS-CoV-2 RNA-Sequences by Interpretable Machine Learning Models\\
2.1\%&Explainable AI reveals key changes in skin microbiome associated with menopause, smoking, aging and skin hydration\\
\bottomrule
    \end{tabular}}
    \caption{Top cited papers in XAI-Biology by papers outside of Biology.}
    \label{tab:xai-bio2non-bio}
\end{table}

\bibliographystyle{ACM-Reference-Format}
\bibliography{sample-base}

\end{document}